# Beyond a Universal Forecasting Selector: Demand-Conditioned Model Selection across Demand Patterns and Horizons

Adolfo González[a]* and Víctor Parada[a]

[a] Department of Computer Engineering and Informatics, Faculty of Engineering, University of Santiago of Chile, Santiago, Chile

* Corresponding author. Email: adolfo.gonzalez.c@usach.cl

ORCID
Adolfo González: https://orcid.org/0009-0008-4452-1285
Víctor Parada: https://orcid.org/0000-0002-8649-5694

E-MAIL
Adolfo González: adolfo.gonzalez.c@usach.cl
Víctor Parada: victor.parada@usach.cl

**Abstract**

Forecasting-model selection remains difficult in heterogeneous demand because the most suitable decision rule may vary with demand structure, data availability, and forecasting horizon. This study examines whether the selector itself should be treated as a context-dependent component of the forecasting process. Five selection mechanisms—RMSSE, ERA, OWA, CCG-AHSC, and CCG-AHSCD—are compared across 24 optimized forecasting models, nine datasets, three training–testing partitions, and horizons from 1 to 12 cycles. Selector performance is evaluated ex post using Global Relative Accuracy (GRA), statistical tests, and a best-attainable-model reference. No selector dominates across all conditions. CCG-AHSC and CCG-AHSCD are more competitive for Smooth demand and several Erratic configurations, whereas OWA and ERA perform better in Intermittent and Lumpy settings. Selector suitability also changes with historical data availability and horizon, supporting a context-dependent rather than universal approach to forecasting-model selection.

## 1. Introduction

Demand forecasting plays a critical role in purchasing, replenishment, inventory management, production, and capacity planning. Forecasting errors can lead to excess inventory, stockouts, capital immobilization, reduced service levels, and operational instability. As a result, forecast quality has consequences that extend beyond statistical accuracy alone (Goltsos et al., 2022; Theodorou et al., 2025; Ulrich et al., 2022). This challenge is especially pronounced in multiproduct environments, where demand series often differ significantly in frequency, variability, intermittency, and sparsity. In multi-horizon operational forecasting, predictive performance is also related to the cumulative demand volume over the planning horizon. Makridakis, Fry, et al. (2022) note that, for applications such as store replenishment and production, cumulative forecast error over the lead time may be more relevant than averaging forecast errors across individual horizons.

The proliferation of statistical, machine-learning, deep-learning, and hybrid forecasting methods has resulted in multiple candidate models for any given demand series. However, empirical evidence demonstrates that no single model consistently outperforms others across all datasets, demand structures, frequencies, or forecasting horizons (Makridakis et al., 2018; Makridakis, Spiliotis, et al., 2022; Petropoulos et al., 2022). Furthermore, model rankings can vary by evaluation metric, as different error measures highlight distinct aspects of predictive performance (Gneiting, 2011; Hyndman & Koehler, 2006; Kolassa, 2020). Therefore, the forecasting challenge extends beyond model training and optimization to include selecting the most appropriate model for each series and decision context.

Automatic model selection offers a scalable solution when numerous products and forecasting methods must be evaluated simultaneously (Garred et al., 2026; Poler & Mula, 2011). Recent research has further shown that model suitability itself is context dependent. Reina-Jiménez et al. (2026), for example, employ time-series meta-features and an interpretable decision-tree meta-learner to recommend forecasting algorithms according to the structural characteristics of the data, reinforcing the view that no single forecasting model is universally appropriate. However, comparatively less attention has been devoted to whether the model-selection mechanism itself should also vary with the forecasting context. This distinction is particularly relevant in heterogeneous demand environments, where Smooth, Intermittent, Erratic, and Lumpy patterns exhibit different structural characteristics (Petropoulos et al., 2022; Syntetos et al., 2005; Syntetos & Boylan, 2005). Accordingly, model selection can be considered a conditioned decision problem in which not only model suitability, but also selector effectiveness, may depend on demand structure, historical data availability, and forecasting horizon.

In this context, the present study evaluates five automatic model-selection mechanisms, each representing a distinct decision logic: Root Mean Squared Scaled Error (RMSSE), Equilibrium Ranking Aggregation (ERA), Overall Weighted Average (OWA), Class-Conditioned GRA with the Adaptive Hybrid Selector (CCG-AHSC), and Class-Conditioned GRA with the Adaptive Hybrid Selector with Class-based Discrimination (CCG-AHSCD). Their comparative suitability is evaluated within a unified experimental framework that includes 24 optimized forecasting models, nine heterogeneous datasets, three training–testing partitions, and forecasting horizons ranging from 1 to 12 cycles. Selector performance is assessed using Global Relative Accuracy (GRA) as an ex post measure of volumetric coherence between forecasted and realized demand, supplemented by statistical comparisons across demand classes and experimental configurations.

Accordingly, this study addresses two related questions: whether any model-selection mechanism consistently outperforms alternative selectors under heterogeneous demand conditions, and how selector suitability varies across combinations of demand class, historical data availability, and forecasting horizon. Rather than proposing another direct forecasting-model recommender, the study examines the model-selection mechanism itself as a distinct decision layer within the forecasting process. An ex post best-model reference is additionally used to establish the attainable performance within the optimized candidate set and to quantify the gap between the model assigned in advance and the best available model identified after future demand is observed. The contribution therefore lies in shifting the analysis from the conventional question of which forecasting model performs best to the higher-level question of which selection mechanism is most appropriate for identifying that model under specific forecasting conditions.

## 2. Literature Review

Demand forecasting is fundamental to purchasing, replenishment, inventory allocation, and resource planning. Forecast quality directly affects operations: overestimation increases inventory costs, obsolescence risk, and capital immobilization, while underestimation leads to stockouts, lost sales, and service deterioration (Kumar et al., 2025; Sepúlveda-Rojas et al., 2015; Sousa et al., 2025; Theodorou et al., 2025; Ulrich et al., 2022). Recent decision-oriented research has further shown that forecast-model misspecification can propagate into subsequent inventory decisions and that predictive objectives may not necessarily align with downstream operational objectives (Y. Zhang et al., 2025). Thus, forecasting performance is significant not only statistically but also operationally, as the chosen forecast determines the demand volume that informs future business decisions.

The proliferation of statistical, machine-learning, deep-learning, and hybrid forecasting methods has expanded the range of models applicable to a given demand series. However, forecasting literature consistently finds that no single model outperforms others across all datasets, demand structures, forecasting horizons, and evaluation criteria (Makridakis et al., 2018; Makridakis, Spiliotis, et al., 2022; Makridakis & Petropoulos, 2020; Petropoulos et al., 2022). Model performance depends on factors such as frequency, seasonality, variability, intermittency, and predictability (Hendry & Pretis, 2023; Syntetos et al., 2005; Syntetos & Boylan, 2005). Therefore, when multiple candidate models exist, the challenge extends beyond fitting or optimizing them to determining which model is most appropriate for each demand series and operational context.

This challenge is particularly acute in multiproduct environments, where manual model selection lacks reproducibility and scalability. Automatic predictive modeling has therefore increasingly incorporated algorithm selection and hyperparameter optimization as explicit components of the modeling process. Salvador et al. (2019), for example, extended the combined algorithm selection and hyperparameter optimization (CASH) paradigm to multicomponent predictive systems, demonstrating that alternative algorithms, preprocessing structures, and parameter configurations can be evaluated within an automated framework. To address this, forecasting research has explored model choice using out-of-sample validation, information criteria, feature-based approaches, meta-learning, expert systems, and multicriteria procedures (Fildes & Petropoulos, 2015; Garred et al., 2026; Tashman, 2000). Automatic selection procedures are especially important when numerous products and forecasting methods require simultaneous evaluation, as expert-driven model construction and hyperparameter tuning become increasingly impractical at scale (Abdallah et al., 2025; Poler & Mula, 2011).

A further complication is that the definition of the "best" model varies with the evaluation criterion. Different error measures yield different model rankings because they prioritize distinct aspects of forecast performance (Badulescu et al., 2021; Hyndman & Koehler, 2006; Makridakis et al., 2018; Makridakis, Spiliotis, et al., 2022). Therefore, model selection should align with the specific forecasting objective and operational application, rather than relying on a single universal accuracy metric (Gneiting, 2011; Kolassa, 2020). This consideration is especially critical in inventory and purchasing, where both pointwise predictive accuracy and the aggregate forecast volume can affect subsequent decisions.

Recent feature-based and meta-learning approaches have addressed this heterogeneity by learning mappings from time-series characteristics to suitable forecasting algorithms. Reina-Jiménez et al. (2026), for example, construct a prescriptive recommendation framework in which statistical and temporal meta-features are used as inputs to an interpretable CART meta-learner that predicts the forecasting algorithm expected to perform best for a given series. This represents a first decision level concerned with identifying the appropriate forecasting model from observable data characteristics. A related but distinct problem concerns the selection mechanism itself: even when several procedures are available for choosing among candidate forecasting models, there is no a priori reason to assume that the same selector will remain equally effective across all forecasting contexts. Accordingly, model-selection research can be viewed as involving two linked decision levels: determining which forecasting model is most appropriate for a series and determining which selection mechanism is most suitable for making that choice.

Demand structure introduces additional complexity at both decision levels. Smooth, Intermittent, Erratic, and Lumpy demand patterns may respond differently not only to identical forecasting models but also to the criteria used to distinguish among those models (Petropoulos et al., 2022; Syntetos et al., 2005; Syntetos & Boylan, 2005). Consequently, conditioning the forecasting decision on demand characteristics may require more than adapting the model itself; the suitability of the selection rule may also vary across demand structures. This motivates examining selector effectiveness explicitly rather than presuming that a single model-selection mechanism is universally appropriate.

Recent research has increasingly addressed forecasting heterogeneity through automated forecast combination and adaptive hybrid modeling. García-Aroca et al. (2024) introduced a principal component analysis–based procedure to rank, select, and combine forecasting methods, thereby reducing reliance on a single error criterion. Similarly, Hammam et al. (2025) developed an adaptive ARIMA–XGBoost framework that modifies the forecasting structure based on the presence of linear and nonlinear demand components. More recently, Zhang et al. (2026) addressed heterogeneous multi-energy demand under limited historical data through a meta-learning graph neural network designed to adapt across diverse user patterns and forecasting tasks. Their results also showed that predictive performance evolves differently as the forecasting horizon increases, providing further evidence that forecast effectiveness is conditioned not only by demand characteristics but also by data availability and horizon length. Collectively, these studies provide further evidence that forecasting performance is contingent on the underlying data characteristics and that fixed modeling strategies may be inadequate in heterogeneous demand environments. Notably, these adaptation efforts have primarily concentrated on the forecasting model, forecast combination, or direct model recommendation. Even recent prescriptive meta-learning approaches, such as Reina-Jiménez et al. (2026), focus on learning which forecasting algorithm should be applied from the characteristics of the series. Comparatively less attention has been devoted to whether the mechanism responsible for selecting among candidate models should itself vary with the forecasting context. This distinction is important because a selector that performs well for one demand pattern, amount of historical information, or forecasting horizon may not retain the same effectiveness under different conditions. Forecasting-model selection can therefore be conceptualized as a conditioned decision problem in which selector suitability depends jointly on demand structure, data availability, and forecasting horizon.

In this context, the central methodological question concerns not only which forecasting model performs best, but also which model-selection procedure most reliably identifies appropriate models across diverse demand conditions and forecasting horizons. This study compares alternative selection mechanisms employing different decision logics, including single-metric, benchmark-relative, multicriteria, and class-conditioned approaches. Their performance is assessed across multiple demand scenarios and forecasting horizons to identify their respective strengths, limitations, and suitable contexts of application. The analysis therefore provides an empirical basis for determining when each selection method is most appropriate, rather than assuming the existence of a universally optimal rule.

## 3. Materials and Methods

The methodological design employed to evaluate forecasting model-selection criteria under structural heterogeneity and horizon-dependent predictive performance is outlined. The section details the datasets, training and testing partitions, forecasting models, performance metrics, inferential methods, ex post volumetric assessment, and selection criteria, with a distinction made between monometric, multicriteria, and adaptive approaches.

### 3.1. Predictive Performance Evaluation Metrics

The evaluation of demand forecasting models necessitates alignment among the forecasting objective, the chosen error metric, and the statistical functional elicited by that metric. Error measures are not interchangeable; squared-error metrics typically correspond to the conditional mean, absolute-error metrics to the conditional median, and percentage-based metrics may induce weighted functionals distinct from both (Gneiting, 2011; Kolassa, 2020). Therefore, each metric in this study is assigned a distinct methodological role, rather than treating all indicators as interchangeable measures of accuracy.

The performance indicators employed include Mean Absolute Error (MAE), Root Mean Squared Error (RMSE), Mean Absolute Scaled Error (MASE), Root Mean Squared Scaled Error (RMSSE),

Mean Absolute Percentage Error (MAPE), symmetric Mean Absolute Percentage Error (sMAPE), coefficient of determination ($R^2$), forecast bias (BIAS), and Global Relative Accuracy (GRA). MAE quantifies the magnitude of absolute errors, RMSE penalizes larger deviations, and MASE and RMSSE offer scale-normalized measures suitable for comparing heterogeneous demand series (Hyndman & Athanasopoulos, 2021; Hyndman & Koehler, 2006; Petropoulos et al., 2022). According to the proposed selection logic, RMSSE is the primary scaled-error criterion, MAE refines absolute-error assessment, sMAPE diagnoses relative-error performance, and BIAS indicates systematic overestimation or underestimation. MAPE and sMAPE are interpreted with caution, as percentage-based metrics may yield undesirable behavior near zero values and may correspond to functionals distinct from those associated with absolute or squared errors (Gneiting, 2011; Hyndman & Koehler, 2006; Kolassa, 2020; Kolassa & Schütz, 2007). $R^2$ is included solely as a descriptive measure of explanatory capacity, while BIAS is pertinent to inventory exposure, replenishment risk, service reliability, and supply chain costs (Doszyń & Dudek, 2024; Sanders & Graman, 2016).

The metrics are therefore organized hierarchically within the adaptive selection framework. RMSSE serves as the principal scale-normalized comparator, MAE refines absolute-error evaluation, sMAPE enhances relative-error assessment, and BIAS is applied for directional diagnosis or as a tiebreaker. This hierarchical structure is especially important because frequent and stable demand allows for multicriteria refinement, while intermittent, irregular, or highly variable demand necessitates more conservative criteria to prevent unstable decisions resulting from sparse observations, zero-demand periods, or extreme values.

### 3.2. Global Relative Accuracy (GRA)

From an operational perspective, demand forecasts specify the volume anticipated for purchase, replenishment, or allocation over a defined future period. In addition to conventional pointwise accuracy measures, evaluating the alignment between total forecasted demand and total realized demand is particularly valuable in inventory and supply-chain contexts, where forecast volume directly influences operational decisions (Kumar et al., 2025; Nguyen, 2026; Sayed et al., 2009; Sepúlveda-Rojas et al., 2015; Ulrich et al., 2022).

The GRA serves as a volumetric performance metric that quantifies the relative agreement between aggregate forecasted and observed demand across the evaluation horizon:

$$GRA = 1 - \frac{\left|\sum_{t=1}^{T} \hat{y}_t - \sum_{t=1}^{T} y_t\right|}{\max\left(\sum_{t=1}^{T} y_t\,, \varepsilon\right)} \tag{1}$$

where $y_t$ denotes observed demand, $\hat{y}_t$ the corresponding forecast, T the number of periods considered, and $\varepsilon = 1$ prevents numerical instability. Values closer to 1 indicate greater agreement between forecasted and realized demand volumes, whereas negative values may occur when the absolute aggregate forecast error exceeds the realized demand volume. Cases with zero aggregate realized demand are excluded.

GRA should be interpreted as a volumetric coherence metric rather than a conventional pointwise accuracy measure. This distinction is significant because aggregate relative measures are not substitutes for statistical accuracy criteria or proper scoring functions for point forecasts (Gneiting, 2011; Kolassa, 2020; Kolassa & Schütz, 2007).

Within the proposed framework, CCG utilizes GRA-derived performance information during the preselection stage. Additionally, GRA is employed in empirical evaluation to assess the ex post volumetric consistency of forecasts produced by the evaluated selectors. Persistent aggregate overestimation or underestimation can result in operational consequences such as excess inventory, immobilized capital, stockouts, lost sales, or service deterioration (Demizu et al., 2023; Goltsos et al., 2022; Nguyen, 2026).

GRA does not explicitly account for holding costs, shortage costs, service-level penalties, replenishment constraints, or production-capacity restrictions. Consequently, it serves as an intermediate indicator of volumetric coherence rather than a comprehensive inventory-cost or production-planning objective.

### 3.3. Model Selectors

This section formalizes the model selection framework adopted in this study, which includes three reference selectors (RMSSE, ERA, and OWA), a class-conditioned preselection procedure (CCG), and two adaptive multicriteria selectors (AHSC and AHSCD). The reference selectors employ distinct decision rules: RMSSE identifies the model with the lowest scaled forecast error, ERA aggregates the ordinal performance of MAE, RMSE, and $R^2$, and OWA assesses candidate models relative to a benchmark using sMAPE and MASE. In contrast, CCG narrows the candidate model space based on the demand class prior to the adaptive selection stage. AHSC and AHSCD then operate on this reduced set using class-specific multicriteria decision rules grounded in Pareto filtering and hierarchical refinement. This class-conditioned approach aligns with recent findings that no single forecasting model is universally optimal across all demand patterns, and that model-selection strategies tailored to demand characteristics can improve the adaptation of forecasting methods to individual series (E et al., 2024). Let $M = \{m_1, m_2, \dots, m_K\}$ denote the set of candidate models evaluated on the same demand series. For each model $m \in M$, the performance information available at the selection stage provides the basis for the criteria formalized below.

#### *3.3.1. RMSSE Model Selector*

The Root Mean Squared Scaled Error (RMSSE) is included as a single-metric comparator to assess predictive accuracy. RMSSE measures forecast error relative to the historical variability of the series, enabling performance comparisons across demand series with varying scales. Its application is well established in large-scale forecasting contexts, such as the M5 competition (Makridakis, Spiliotis, et al., 2022). For a candidate model, RMSSE is defined as:

$$RMSSE_m = \sqrt{\frac{\frac{1}{H}\sum_{h=1}^{H}\left(y_{n+h} - \hat{y}_{m,n+h}\right)^2}{\frac{1}{n-1}\sum_{t=2}^{n}(y_t - y_{t-1})^2}} \quad (2)$$

where $H$ denotes the forecast horizon, $n$ is the number of training observations, $y_{n+h}$ is the observed value at horizon $h$, and $\hat{y}_{m,n+h}$ is the forecast generated by model $m$. The denominator provides the scaling factor based on consecutive changes in the training series. For each series $i$, the selected model is the candidate with the lowest RMSSE:

$$m^*_{i,RMSSE} = \underset{m \in \mathcal{M}}{\operatorname{argmin}}\ RMSSE_{i,m} \tag{3}$$

where $\mathcal{M}$ denotes the set of candidate models. Thus, the selector applies a uniform and reproducible minimum-error rule across all evaluated series.

### *3.3.2. Equilibrium Ranking Aggregation Selector*

The Equilibrium Ranking Aggregation (ERA) selector is included as a multicriteria ranking-based comparator. It evaluates candidate forecasting models through the joint ordinal performance of three accuracy indicators: MAE, RMSE, and the coefficient of determination ($R^2$). This type of multicriteria assessment is consistent with demand forecasting studies that relate predictive accuracy to operational performance in inventory management (Kumar et al., 2025).

For each candidate model $m$, MAE and RMSE are ranked in ascending order, whereas $R^2$ is ranked in descending order. The ERA score is defined as:

$$ERA_m = \text{Rank}(MAE_m) + \text{Rank}(RMSE_m) + \text{Rank}(R^2_m) \tag{4}$$

where lower values indicate better aggregate ordinal performance. The selected model is therefore:

$$m^*_{ERA} = \underset{m \in \mathcal{M}}{\operatorname{argmin}}\ ERA_m \tag{5}$$

where $\mathcal{M}$ denotes the set of candidate models. Thus, ERA applies a uniform multicriteria ranking rule and selects the model with the lowest aggregate rank.

### *3.3.3. Overall Weighted Average*

The Overall Weighted Average (OWA) criterion serves as a benchmark-relative comparator. First introduced as an evaluation metric in the M4 Competition, OWA simultaneously incorporates two scale-independent accuracy measures, MASE and sMAPE, in relation to the Naïve2 benchmark (Makridakis et al., 2018; Makridakis & Petropoulos, 2020). The Naïve2 method offers a seasonally adjusted naïve reference and has been extensively utilized as a benchmark in forecasting competitions and related studies. This benchmark-relative framework has also been implemented in automated model selection for supply chain forecasting (Garred et al., 2026). For each candidate model m, the OWA score is calculated as follows:

$$OWA_m = \frac{1}{2}\left(\frac{MASE_m}{MASE_b} + \frac{sMAPE_m}{sMAPE_b}\right) \tag{6}$$

where MASE_b and sMAPE_b denote the corresponding errors of the benchmark model. Lower OWA values indicate better performance relative to the benchmark. The selected model is therefore:

$$\mathrm{m}^*_{\mathrm{OWA}} = \arg\min_{\mathrm{m} \in \boldsymbol{\mathcal{M}}} \mathrm{OWA_m} \tag{7}$$

where $\boldsymbol{\mathcal{M}}$ denotes the set of candidate models. Thus, OWA applies a uniform benchmark-relative selection rule and selects the model with the lowest combined relative error.

### *3.3.4. Class-Conditioned GRA (CCG)*

The CCG is an adaptive preselection procedure intended to reduce the initial set of forecasting models for each demand series or stock-keeping unit (SKU) prior to applying a subsequent selection mechanism. The procedure aims to identify models that demonstrate stronger evidence of favorable performance during training while explicitly considering the underlying demand structure.

Unlike a uniform filtering procedure, CCG applies differentiated evaluation rules for Smooth, Intermittent, Erratic, and Lumpy demand classes. Consequently, both the performance characteristics used in the evaluation and the number of models retained depend on the demand class assigned to the series. For a given SKU, let $M_{valid} \subseteq M$ denote the subset of candidate models with valid forecasting results, and let $N = |M_{valid}|$ denote the number of valid models. The characteristics used by CCG are transformed into relative quality measures. For a given characteristic $x$, let $r_{x(m)}$ denote the rank achieved by model $m$, where rank 1 corresponds to the best-performing model. The relative quality associated with characteristic $x$ is defined as:

$$q_{x(m)} = 1 - \frac{r_{x(m)} - 1}{N - 1} \tag{8}$$

such that higher values of $q_{x(m)}$ indicate better relative performance. For Smooth demand, CCG jointly considers the model's initial performance and its worst observed ranking:

$$S_{Smooth(m)} = \frac{q_{first(m)} + q_{worstRank(m)}}{2} \tag{9}$$

Here, $q_{first(m)}$ represents the relative quality of the initial performance, whereas $q_{worstRank(m)}$ reflects the quality associated with the worst ranking reached by the model. For Intermittent demand, the criterion primarily emphasizes the model's ability to remain among the best-performing alternatives:

$$S_{base(m)} = 0.999\, Top3Freq(m) + 0.001\, q_{slope(m)} \tag{10}$$

$$S_{Intermittent(m)} = 0.90\, S_{base(m)} + 0.10\, q_{recentSlope4(m)} \tag{11}$$

In this formulation, $Top3Freq(m)$ represents the frequency with which the model ranks among the three best alternatives, $q_{slope(m)}$ characterizes its overall performance trend, and $q_{recentSlope4(m)}$ captures its recent evolution. For Erratic demand, a broader combination of characteristics is used:

$$S_{(GRA)(m)} = \frac{Top5Freq(m) + Top8Freq(m) + q_{rankSlope(m)}}{3} \tag{12}$$

$$S_{base(m)} = 0.85\, S_{(GRA)(m)} + 0.15\, q_{sMAPE(m)} \tag{13}$$

$$S_{Erratic(m)} = 0.90\, S_{base(m)} + 0.10\, q_{acceleration3(m)} \tag{14}$$

Here, $Top5Freq(m)$ and $Top8Freq(m)$ measure how frequently the model remains among the five and eight best alternatives, respectively; $q_{rankSlope(m)}$ represents the evolution of its ranking, $q_{sMAPE(m)}$ its relative quality according to sMAPE, and $q_{acceleration3(m)}$ the recent change in its performance trend. For Lumpy demand, CCG considers the strongest evidence obtained from different dimensions of model behavior:

$$S_{(GRA)(m)} = \max\left(q_{mean(m)}, q_{first(m)}, q_{recentRank3(m)}\right) \quad (15)$$

$$S_{Lumpy(m)} = 0.55\, S_{(GRA)(m)} + 0.45\, q_{MASE(m)} \quad (16)$$

In this case, $q_{mean(m)}$ represents the relative quality of the model's average performance, $q_{first(m)}$ its initial performance, $q_{recentRank3(m)}$ its recent ranking behavior, and $q_{MASE(m)}$ its relative quality according to MASE. Once the class-specific score $S_{c(m)}$ is computed, the valid models are ranked in descending order and only the $K_c$ best candidates are retained:

$$\mathcal{M}^{c}_{CCG} = TopK\left(S_{c(m)}, K_c\right) \quad (17)$$

Thus, CCG performs a class-conditioned reduction of the original candidate set, producing a reduced subset

$$\mathcal{M}^{c}_{CCG} \subseteq \mathcal{M}_{valid} \quad (18)$$

which is subsequently passed to the final model selection procedure, such as AHSC or AHSCD. In this way, CCG acts as a class-dependent screening mechanism rather than as a final selector, reducing the decision space before the definitive model selection stage.

The class-specific weighting coefficients and candidate-retention values used by CCG were derived from the analysis of preliminary experimental results conducted prior to the final comparative evaluation. These preliminary analyses were used to identify the relative importance of the training-performance characteristics associated with each demand class and to determine the size of the candidate subset retained for subsequent selection. Importantly, these parameters were fixed before the final evaluation and were not adjusted using the test-period outcomes reported in this study. Once specified, the weighting coefficients and $K_c$ values remained unchanged across all datasets, training–testing partitions, and forecasting horizons. This fixed-parameter design was adopted to prevent adaptation of the selector rules to the observed test results and to preserve comparability across experimental configurations.

### *3.3.5. Class-Conditioned Adaptive Hybrid Selector (AHSC)*

The AHSC is a hierarchical multicriteria procedure developed to identify, for each demand series or stock-keeping unit (SKU), the forecasting model that achieves the optimal balance across multiple performance criteria. AHSC operates on the set of models previously filtered by the CCG procedure, ensuring that only the retained candidates advance to subsequent selection stages. This method employs multiobjective optimization, where Pareto-optimal solutions maintain alternatives that are not simultaneously dominated across the considered criteria (Sinha et al., 2015; Williams et al., 2019).

Following CCG filtering, AHSC implements a sequential strategy. Initially, non-dominated models are retained in the first Pareto front. Subsequently, a Minimax criterion identifies models with the lowest worst-case relative performance compared to a benchmark model. If multiple candidates remain, a refinement metric is applied to provide a robust decision rule against unfavorable outcomes in any primary criterion (Petrov, 2024). The first Pareto front is defined as:

$$P_1 = \{m \in M_{CCG}^{c} : \nexists\, m' \in M_{CCG}^{c}\, such\ that\ m' \prec m\} \quad (19)$$

The Minimax candidate set is obtained as:

$$C = \arg\min_{m \in P_1} \left[ \max_{j=1,2} \left( \frac{f_{j(m)}}{f_{j(m_b)}} \right) \right] \quad (20)$$

The final model is selected using the refinement criterion:

$$m^{*} = \arg\min_{m \in C} r(m) \quad (21)$$

where $M_{CCG}^{c}$ denotes the set of candidate models retained by CCG for demand class $c$, $P_1$ the first Pareto front, $f_1$ and $f_2$ and the primary evaluation metrics, $m_b$ the benchmark model used for relative normalization, $r(m)$ the refinement metric, and $m^{*}$ the selected model. The metrics used at each stage are specified according to the demand class, allowing the selection rule to reflect differences among Smooth, Erratic, Intermittent, and Lumpy demand patterns (Sarlo et al., 2023).

#### ***3.3.6. Adaptive Hybrid Selector with Class-based Discrimination (AHSCD)***

The AHSCD is a hierarchical multicriteria procedure developed to identify, for each demand series or stock-keeping unit (SKU), the forecasting model that achieves the optimal balance across multiple performance criteria. AHSCD is applied to the set of models previously filtered by the CCG method, ensuring that only the retained candidates proceed to subsequent selection stages. The procedure is grounded in multiobjective optimization and post-Pareto selection, wherein the non-dominated set is initially identified and the final alternative is subsequently determined using additional decision criteria (Carrillo et al., 2011; Williams et al., 2019).

AHSCD employs a sequential strategy following CCG filtering. Initially, non-dominated models are retained in the first Pareto front. Subsequently, a class-specific discriminating metric further reduces this set by retaining the alternatives with the lowest discriminator value. If multiple candidates remain, a refinement metric is applied to determine the final selection. This approach aligns with post-Pareto decision procedures, which distinguish between the generation of the Pareto front and the subsequent selection of a preferred solution using supplementary information (Carrillo et al., 2011; Williams et al., 2019).

Using the candidate set $M_{CCG}^{c}$ previously retained by CCG, the first Pareto front is defined as:

$$P_1 = \{m \in M_{CCG}^{c} : \neg\exists\, m' \in M_{CCG}^{c}\, such\ that\ m' \prec m\} \quad (22)$$

The candidate set obtained using the class-specific discriminating metric is defined as:

$$D = \arg\min_{m \in P_1} d_{c(m)} \quad (23)$$

where $d_{c(m)}$ denotes the discriminating criterion associated with the corresponding demand class. Finally, the selected model is determined using the refinement criterion:

$$m^* = \arg\min_{m \in D} r_{c(m)} \tag{24}$$

where $M_{CCG}^{c}$ denotes the set of candidate models retained by CCG for demand class $c$, $P_1$ the first Pareto front, $d_{c(m)}$ the class-conditioned discriminating metric, $D$ the set of candidates retained after the discrimination stage, $r_{c(m)}$ the refinement metric, and $m^*$ the selected model. The metrics used in the Pareto, discrimination, and refinement stages are specified according to the demand class, allowing the selection rule to account for differences among Smooth, Erratic, Intermittent, and Lumpy demand patterns (Sarlo et al., 2023).

### 3.4. Backtesting Strategy

For each demand series, the final observations are allocated as an out-of-sample evaluation segment, with all prior observations utilized for model estimation. The forecasting horizon is defined according to the intrinsic frequency of each series, such as daily, weekly, monthly, or annual, thereby maintaining alignment with the relevant operational planning unit.

A fixed-origin multi-step backtesting strategy is implemented, wherein candidate models are estimated solely on observations preceding the evaluation segment and subsequently generate forecasts for the reserved cycles. The resulting predictions are compared with the realized values to calculate out-of-sample performance measures. This approach preserves temporal ordering, prevents information leakage, and aligns with established practices for multi-step forecast evaluation (Hyndman & Koehler, 2006; Kourentzes et al., 2014; Semenoglou et al., 2021; Tashman, 2000), as well as with demand forecasting applications that support inventory and operational planning decisions (Armstrong, 2001; Makridakis, Spiliotis, et al., 2022).

### 3.5. Experimental Design

The experimental framework assesses automatic model selection in heterogeneous demand forecasting environments by systematically comparing five selection mechanisms within a unified out-of-sample evaluation context: RMSSE, ERA, OWA, CCG-AHSC, and CCG-AHSCD. RMSSE employs a single-metric selection rule based on scaled forecast error, ERA utilizes a multicriteria ranking-aggregation approach, and OWA functions as a benchmark-relative selector using MASE and sMAPE. In contrast, CCG-AHSC and CCG-AHSCD represent demand-class-conditioned adaptive pipelines, where CCG initially reduces the candidate model set based on the structural characteristics of the demand series, followed by AHSC or AHSCD conducting the final multicriteria selection. The workflow standardizes candidate models, temporal validation schemes, performance metrics, and ex post GRA assessment of volumetric coherence across all selection mechanisms. Figure 1 illustrates the experimental design, which comprises 10 stages grouped into four blocks: data preparation, model estimation, selector-based assignment, and ex post forecast assessment.

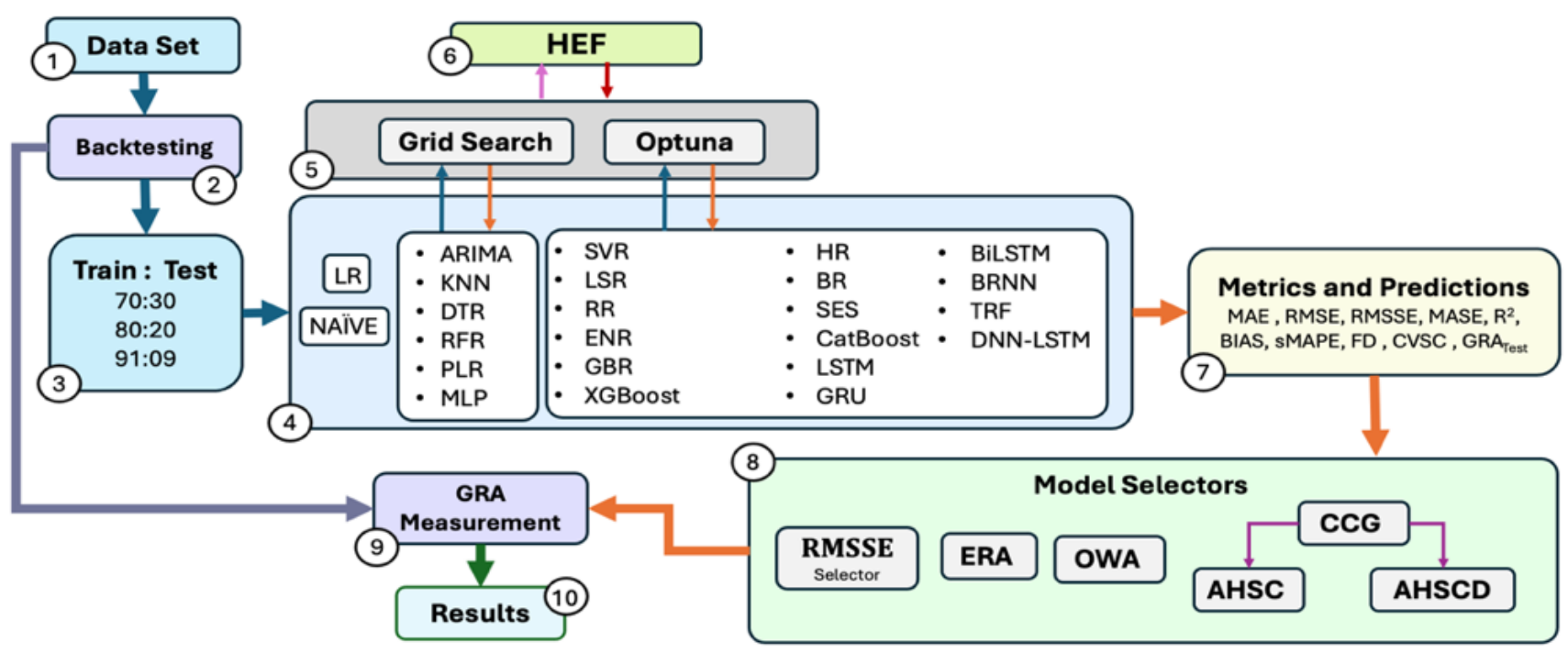

**Figure 1.** Experimental Framework.

### *3.5.1. Stage 1: Dataset.*

The empirical analysis utilized nine established forecasting datasets that encompass a range of demand frequencies, sample sizes, and demand-pattern structures: Walmart (Yasser, 2021), M5 (Makridakis, Spiliotis, et al., 2022), BRAF (de Haan, 2021a), M4 (Makridakis et al., 2018), Beer (Kilts Center for Marketing, n.d.), MAN (de Haan, 2021b), M3 (Makridakis & Hibon, 2000), Analgesics (Kilts Center for Marketing, n.d.), and Cheeses (Kilts Center for Marketing, n.d.).

To characterize demand heterogeneity within the experimental sample, each time series was categorized as Smooth, Intermittent, Erratic, or Lumpy based on the demand-pattern taxonomy proposed by Syntetos et al. (2005). Table 1 presents the main characteristics of the datasets and the distribution of series across these demand classes.

**Table 1.** Datasets used in the experiments.

| Dataset | Frequency | # Time Series | # Observations | Classes | | | |
|---|---|---|---|---|---|---|---|
| | | | | # Smooth | # Intermittent | # Erratic | # Lumpy |
| Walmart | Weekly | 45 | 4,680 | 45 | 0 | 0 | 0 |
| M5 | Daily | 643 | 231,480 | 60 | 489 | 15 | 79 |
| BRAF | Monthly | 586 | 28,128 | 0 | 373 | 0 | 213 |
| M4 | Weekly | 294 | 29,780 | 293 | 0 | 1 | 0 |
| Beer | Weekly | 648 | 76,464 | 47 | 392 | 45 | 164 |
| MAN | Weekly | 850 | 30,600 | 3 | 386 | 41 | 420 |
| M3 | Monthly | 1,428 | 51,408 | 1,414 | 0 | 14 | 0 |
| Analgesics | Weekly | 437 | 51,566 | 169 | 143 | 40 | 85 |
| Cheeses | Weekly | 402 | 47,436 | 172 | 97 | 73 | 60 |

### *3.5.2. Stage 2: Backtesting.*

A fixed-origin temporal backtesting scheme was adopted to preserve the chronological structure of each series and prevent leakage of future information (Hyndman & Koehler, 2006; Tashman, 2000). For all datasets, the future evaluation horizon was set to $H = 12$ cycles, with the notion of cycle adapted to the intrinsic frequency of each dataset. Accordingly, the evaluation horizon is 12 weeks for the weekly series, 12 months for the monthly series, and 12 daily periods for the daily series. This design ensures comparability across datasets while remaining consistent with the operational planning unit of each series.

### *3.5.3. Stage 3: Training and test separation.*

Each sampled dataset was evaluated using three training and testing configurations: 91:9, 80:20, and 70:30. These partitioning schemes facilitate analysis of selector behavior under varying levels of historical data availability while maintaining a consistent evaluation protocol across all experiments.

#### *3.5.4. Stage 4: Demand forecasting models.*

The candidate forecasting set consists of 24 models, including statistical, machine learning, ensemble-based, and deep learning approaches. These models are categorized by optimization strategy into Exhaustive Search (ES) and Search in Continuous Space (SCS). Appendix A, Table A1, provides the complete list of models, their acronyms, methodological families, and optimization groups.

#### *3.5.5. Stage 5: Optimizers.*

Two optimization strategies were selected according to the model structure. Grid Search was implemented for ES models to ensure exhaustive exploration of the predefined discrete parameter space. Optuna was employed for SCS models, facilitating efficient hyperparameter optimization in continuous spaces through Bayesian search and pruning (Akiba et al., 2019).

#### *3.5.6. Stage 6: HEF.*

All model parameters, hyperparameters, and fixed configuration settings were calibrated or defined using the Hierarchical Evaluation Function (HEF) (González & Parada, 2026) as a unified optimization criterion. HEF ensured methodological consistency across models by integrating explanatory power, average error magnitude, and sensitivity to large errors within a single evaluation structure. This approach reduces the risk that subsequent selector comparisons reflect differences in optimization criteria rather than genuine differences in predictive performance.

Appendix A, Table A2, reports the complete model-specific search spaces and fixed settings used during training and optimization. The optimization process employed two distinct strategies. For models in the Search in Continuous Space (SCS) group, Optuna was applied with a fixed budget of 21 trials per training and optimization cycle. For models in the Exhaustive Search (ES) group, Grid Search systematically evaluated all combinations within the predefined discrete search spaces.

#### *3.5.7. Stage 7: Metrics and predictions.*

After training under the designated partition scheme, the models produced multi-step forecasts for horizons h = 1 to h = 12. Subsequently, MAE, RMSE, RMSSE, MASE, $R^2$, BIAS, sMAPE, GRATest, FD, and CVSC were calculated for the corresponding evaluation segments. The notation GRATest denotes the GRA information generated from the training-side model-evaluation process and used as input to CCG during candidate screening. Despite its variable name, $GRA_{Test}$ does not refer to the reserved future test period used for final selector assessment. It is therefore distinct from the ex post GRA subsequently computed after model selection using the reserved future observations. Only $GRA_{Test}$, derived prior to the ex post evaluation stage, was available to CCG; the ex post GRA did not enter the model-selection process.

#### *3.5.8. Stage 8: Model selectors.*

Based on the performance information generated in the preceding stages, five model-selection mechanisms were evaluated for each demand series: RMSSE, ERA, OWA, CCG-AHSC, and CCG-AHSCD. These mechanisms encompass a range of decision strategies, including single-metric, benchmark-relative, and class-conditioned multicriteria procedures.

The complete candidate pool consists of K = 24 forecasting models. RMSSE serves as a monometric reference by selecting the candidate model with the lowest scaled error. ERA aggregates ordinal rankings derived from MAE, RMSE, and $R^2$. In contrast, OWA evaluates each candidate relative to a benchmark model using MASE and sMAPE.

For the adaptive procedures, CCG is first applied as a class-conditioned preselection stage. Based on the demand class, CCG reduces the initial candidate set by retaining the $K_c$ best-ranked valid models according to the class-specific scoring rules defined previously. The resulting candidate set $\mathcal{M}_{CCG}^{c}$ is then passed to AHSC or AHSCD for final model selection. AHSC applies Pareto filtering followed by a Minimax decision rule and a refinement stage, whereas AHSCD applies Pareto filtering followed by a class-specific discrimination criterion and subsequent refinement. Therefore, CCG-AHSC and CCG-AHSCD constitute sequential selection pipelines.

For AHSC and AHSCD, the metrics applied at each decision stage are determined by the demand class assigned to each series. The class-specific configurations used in the experiments are summarized in Table 2 and remain fixed across all datasets and training–testing partitions.

**Table 2.** Class-dependent metric configuration of AHSC and AHSCD.

| Selector | Demand class | Pareto metrics | Intermediate criterion | Refinement metric |
|---|---|---|---|---|
| AHSC | Smooth | sMAPE, BIAS | Minimax | \| BIAS \| |
| AHSC | Intermittent | RMSE, MAPE | Minimax | \| BIAS \| |
| AHSC | Erratic | RMSE, MASE | Minimax | MAPE |
| AHSC | Lumpy | RMSSE, BIAS | Minimax | \| BIAS \| |
| AHSCD | Smooth | MASE, BIAS | sMAPE | \| BIAS \| |
| AHSCD | Intermittent | MAPE, sMAPE | MASE | \| BIAS \| |
| AHSCD | Erratic | MAE, sMAPE | RMSSE | \| BIAS \| |
| AHSCD | Lumpy | MAPE, sMAPE | \| BIAS \| | MAE |

Algorithms 1–6 provide an overview of the operational implementation of the five model-selection mechanisms. Six algorithms are included because CCG serves as a shared preselection stage for the two adaptive pipelines, CCG-AHSC and CCG-AHSCD. As model estimation, hyperparameter optimization, forecast generation, and metric computation are completed in earlier stages, the selector algorithms utilize the available performance information directly.

**Algorithm 1.** RMSSE Model Selector

```
Input:
  Candidate model set M
  Validation RMSSE values

Output:
  Best model m*

BEGIN
  best_RMSSE ← ∞
  m* ← NULL

  FOR each model m IN M DO
    Obtain RMSSE_m

    IF RMSSE_m < best_RMSSE THEN
      best_RMSSE ← RMSSE_m
```

```
        m* ← m
      END IF

   END FOR

   RETURN m*

END
```

**Algorithm 2.** ERA Model Selector

---

```
Input:
  Candidate model set M
  Validation metrics MAE, RMSE, and R^2

Output:
  Best model m*

BEGIN
   Rank models by MAE in ascending order
   Rank models by RMSE in ascending order
   Rank models by R^2 in descending order

   FOR each model m IN M DO
      Compute ERA_score_m
      as the sum of its three metric ranks
   END FOR

   Select the model with the lowest ERA_score

   RETURN m*

END
```

**Algorithm 3.** OWA Model Selector

---

```
Input:
  Candidate model set M
  Validation metrics MASE and sMAPE
  Benchmark model b

Output:
  Best model m*

BEGIN
   Obtain MASE_b and sMAPE_b from benchmark model b
   best_OWA ← ∞
   m* ← NULL

   FOR each model m IN M DO
      Obtain MASE_m
      Obtain sMAPE_m
      OWA_m ← 0.5 × ( MASE_m / MASE_b + sMAPE_m / sMAPE_b  )

      IF OWA_m < best_OWA THEN
         best_OWA ← OWA_m
         m* ← m
      END IF

   END FOR
   RETURN m*
END
```

**Algorithm 4.** CCG Candidate Screening

---

```
Input:
  Candidate model set M
  Demand class c
  Training GRA information

Output:
  Filtered candidate set Mc

BEGIN
```

```
  FOR each valid model m IN M DO
     Compute GRA-based performance characteristics
     Transform required characteristics
     into relative quality scores q ∈ [0,1]
  END FOR

  IF c = Smooth THEN
     Kc ← 14
     FOR each valid model m IN M DO
        Score(m) ← (q_first(m) + q_worstRank(m)) / 2
     END FOR
  ELSE IF c = Intermittent THEN
     Kc ← 16
     FOR each valid model m IN M DO
        Score_base(m) ← 0.999 × Top3Freq(m) + 0.001 × q_slope(m)
        Score(m) ← 0.90 × Score_base(m) + 0.10 × q_recentSlope4(m)
     END FOR
  ELSE IF c = Erratic THEN
     Kc ← 12
     FOR each valid model m IN M DO
        Score_GRA(m) ← [Top5Freq(m) + Top8Freq(m) + q_rankSlope(m)] / 3
        Score_base(m) ← 0.85 × Score_GRA(m) + 0.15 × q_sMAPE(m)
        Score(m) ← 0.90 × Score_base(m) + 0.10 × q_acceleration3(m)
     END FOR
  ELSE IF c = Lumpy THEN
     Kc ← 13
     FOR each valid model m IN M DO
        Score_GRA(m) ← max(q_mean(m), q_first(m), q_recentRank3(m) )
        Score(m) ← 0.55 × Score_GRA(m) + 0.45 × q_MASE(m)
     END FOR
  END IF

  Rank all valid models in descending order according to Score(m)
  M^c_CCG ← first Kc models in the ranking

  RETURN M^c_CCG
END
```

**Algorithm 5.** AHSC Model Selector

---

```
Input:
  Filtered candidate model set Mc from CCG
  Demand class c
  Class-specific primary metrics f1 and f2
  Benchmark model b
  Refinement metric r

Output:
  Best model m*

BEGIN
   Select f1, f2, and r according to demand class c

   Compute the first Pareto front P1 from M^c_CCG
   using f1 and f2

   best_minimax ← ∞
   CandidateSet ← empty

   FOR each model m IN P1 DO

      Compute relative performance of f1
      with respect to benchmark model b

      Compute relative performance of f2
      with respect to benchmark model b

      minimax_m ← maximum relative performance across f1 and f2

      IF minimax_m < best_minimax THEN
         best_minimax ← minimax_m
         CandidateSet ← {m}
```

```
    ELSE IF minimax_m = best_minimax THEN
      Add m to CandidateSet
    END IF

  END FOR

  IF CandidateSet contains more than one model THEN
    Select the model with the best refinement metric r
  ELSE
    Select the only model in CandidateSet
  END IF

  RETURN m*

END
```

**Algorithm 6.** AHSCD Model Selector

---

```
Input:
  Filtered candidate model set Mc from CCG
  Demand class c
  Class-specific Pareto metrics f1 and f2
  Class-specific discriminator metric d
  Class-specific refinement metric r

Output:
  Best model m*

BEGIN
  Select f1, f2, d, and r according to demand class c
  Compute the first Pareto front P1 from M^c_CCG
  using f1 and f2

  best_d ← ∞
  CandidateSet ← empty

  FOR each model m IN P1 DO
    Compute discriminator value d_m

    IF d_m < best_d THEN
      best_d ← d_m
      CandidateSet ← {m}

    ELSE IF d_m = best_d THEN
      Add m to CandidateSet
    END IF
  END FOR

  IF CandidateSet contains more than one model THEN
    Select the model with the best refinement metric r
  ELSE
    Select the only model in CandidateSet
  END IF

  RETURN m*

END
```

From a computational perspective, the selector layer introduces a relatively small additional cost because it operates after model estimation, hyperparameter optimization, forecast generation, and metric computation. Let $K$ denote the number of forecasting models in the initial candidate set, with $K$=24 in the present experimental design. RMSSE and OWA require $O(K)$ operations, whereas ERA requires $O(K \log K)$ primarily because the candidate models must be ranked according to their evaluation metrics. CCG computes class-specific performance characteristics and ranks the valid models before retaining the reduced candidate set, resulting primarily in $O(K \log K)$ complexity.

AHSC and AHSCD subsequently operate only on $\mathcal{M}^{c}_{CCG}$, rather than on the complete set of 24 models. Let $K_c$ denote the number of candidates retained by CCG for demand class c. When the first Pareto front is obtained through pairwise dominance comparisons, AHSC and AHSCD have a worst-case complexity of $O(K_c^2)$. The subsequent Minimax, discrimination, and refinement stages require at most linear operations over the Pareto-efficient subset and therefore do not change the overall worst-case order. Given the relatively small candidate sets retained by CCG, the computational cost of the selector layer remains minor relative to model training and hyperparameter optimization.

Accordingly, the five mechanisms should be interpreted as alternative decision strategies originating from the same pool of 24 previously trained forecasting models. RMSSE represents a single-metric minimum-error strategy, ERA an ordinal multicriteria aggregation strategy, and OWA a benchmark-relative strategy. In contrast, CCG-AHSC and CCG-AHSCD constitute sequential class-conditioned pipelines in which CCG first reduces the candidate space and AHSC or AHSCD subsequently performs the final multicriteria selection. This common experimental structure allows differences in forecasting performance to be attributed primarily to the model-selection logic rather than to differences in model training, hyperparameter optimization, or forecast generation.

#### *3.5.9. Stage 9: GRA measurement.*

Following the model assignment stage, predictions generated by the selected models were compared ex post with the observed future demand values reserved during the backtesting phase. This comparison enabled calculation of the Global Relative Accuracy (GRA) indicator, which serves as an ex post aggregate measure of volumetric coherence between forecasted and realized demand. GRA was computed for all horizons from 1 to 12 and for each training–testing partition scheme, facilitating comparison of selector performance across various forecasting horizons.

Statistical comparisons among selectors were conducted separately for each demand class and training–testing partition. Because the same experimental conditions were evaluated across all five selectors, the comparisons were treated as paired. The Shapiro–Wilk test was first used to assess normality of the GRA distributions. Given the non-normal behavior observed in several configurations, nonparametric procedures were adopted for inferential comparison. The Friedman test was applied as the global test for differences among the five selectors, using a significance level of $\alpha = 0.05$. When the Friedman test indicated a statistically significant global difference, pairwise Wilcoxon signed-rank tests were subsequently performed. The resulting pairwise p-values were adjusted using the Holm procedure to control the family-wise error rate. When the global Friedman test was not statistically significant, differences among selector rankings were interpreted descriptively and no claim of pairwise superiority was made.

#### *3.5.10. Stage 10: Results.*

The final stage compares the performance of RMSSE, ERA, OWA, CCG-AHSC, and CCG-AHSCD across datasets, demand classes, forecasting horizons, and training–testing partition schemes. The analysis integrates GRA-based descriptive summaries, distributional plots, inferential comparisons based on the statistical procedure defined above, and aggregated selector rankings across horizons.

As part of this stage, an ex post reference model is identified independently for each SKU by considering all valid and optimized forecasting models applied to that series. Let $\mathrm{M_i}$ denote the set of models available for SKU i. For each model $m \in M_i$, its volumetric performance trajectory over the complete $H = 12$ future cycles is considered.

$$G_m = \left(GRA_{m,1}, GRA_{m,2}, \dots, GRA_{m,H}\right), \;\; H = 12 \tag{25}$$

From this trajectory, the mean and median GRA, the frequency with which the model reaches the first position, the number of exclusive wins, its average ranking across horizons, and two dispersion measures, IQR and MAD, are computed. The mean GRA and average ranking are defined as follows.

$$G\bar{R}\bar{A}_m = \frac{1}{H}\sum_{h=1}^{H} GRA_{m,h} \tag{26}$$

$$\bar{R}_m = \frac{1}{H}\sum_{h=1}^{H} R_{m,h} \tag{27}$$

where $R_{m,h}$ denotes the ranking position of model $m$ at horizon $h$, with higher GRA values corresponding to better positions.

Model comparison follows a hierarchical criterion. Priority is given, successively, to a higher mean GRA, a higher median GRA, a greater frequency of first-place occurrences, a greater number of exclusive wins, a lower average ranking, and lower dispersion according to IQR and MAD. This decision rule can be represented by the criterion vector:

$$V_m = (G\bar{R}\bar{A}_m, Median(G_m), P_m, W_m, -\bar{R}_m, -IQR_m, -MAD_m) \tag{28}$$

where $P_m$ denotes the number of horizons in which model $m$ reaches the first position, and $W_m$ denotes the number of horizons in which it achieves an exclusive win. The best reference model for SKU i is then obtained through lexicographic comparison.

$$m_i^* = \arg\max_{m \in M_i} V_m \tag{29}$$

Accordingly, the mean GRA is considered first, while the remaining criteria are applied sequentially when additional discrimination among candidate models is required. Two models are regarded as tied only when their complete GRA trajectories are equivalent within the predefined numerical tolerance.

For each SKU, this procedure identifies the model, among those effectively applied, that demonstrates the best overall performance across the entire 12-cycle horizon. The resulting reference represents the maximum performance achievable within the set of optimized models available for that SKU. If a selection procedure could identify the same model prior to observing future outcomes, it would achieve the optimal result within the considered model space. The ex post reference therefore serves as an empirical upper bound for assessing how closely the evaluated selectors approximate the best attainable choice.

The analyses evaluate the robustness, consistency, and relative performance of RMSSE, ERA, OWA, CCG-AHSC, and CCG-AHSCD under heterogeneous demand conditions. Special attention is devoted to their behavior across different forecasting horizons and demand classes, enabling identification of not only the best-performing selector overall, but also the specific conditions under which each selection strategy yields the most favorable results.

### 3.6. Experimental Protocol

A standardized experimental protocol was established to evaluate five model-selection mechanisms across nine datasets and three training–testing partition schemes (91:9, 80:20, and 70:30), resulting in 27 distinct experimental configurations. The methodological workflow described in previous stages was consistently applied to all configurations. This process included candidate-model training and optimization, multi-step forecast generation, performance-metric computation, and application of RMSSE, ERA, OWA, CCG-AHSC, and CCG-AHSCD. Forecasts generated by the selected models were subsequently evaluated ex post using GRA over horizons ranging from 1 to 12. For each SKU, an ex post reference was identified from the set of valid and optimized models to establish an empirical upper bound for comparing selector performance. The protocol was consistently maintained across all datasets and partition schemes to ensure comparability and reproducibility.

### 3.7. Hardware and Software

Forecasting models were implemented in Python within a standard scientific computing and machine learning environment. Statistical time-series models utilized Statsmodels (Seabold & Perktold, 2010), while regression and machine learning models employed Scikit-learn (Pedregosa et al., 2011). Gradient-boosting models were developed using XGBoost and CatBoost (Chen & Guestrin, 2016; Prokhorenkova et al., 2018). Neural network architectures were implemented with TensorFlow, Keras, and PyTorch (Abadi et al., 2016; Chollet, 2015; Paszke et al., 2019). Data manipulation and analysis were performed using Pandas, NumPy, and SciPy (Harris et al., 2020; The pandas development team, 2026; Virtanen et al., 2020). Visualization was conducted with Matplotlib and Seaborn (Hunter, 2007; Waskom, 2021). Hyperparameter optimization was conducted using Grid Search or Optuna, selected according to the structural characteristics of each model (Akiba et al., 2019).

Computational experiments were conducted using Jupyter Notebooks within a Conda environment, utilizing Python version 3.13.9 on a Linux Ubuntu operating system. Experiments were executed on a 64-bit x86 architecture with an AMD Ryzen AI 9 HX 370 processor, featuring 12 physical cores, 24 logical threads, and 32 GB of LPDDR5X RAM. The system included an NVIDIA GeForce RTX 5070 Laptop GPU with 8 GB of GDDR7 memory, integrated AMD Radeon Graphics, and an AMD XDNA neural processing unit capable of up to 50 TOPS. Storage was provided by a 2 TB M.2 NVMe PCIe 4.0 SSD. All experiments were performed within the same software environment to ensure consistency and reproducibility.

## 4. Results

The experimental analysis evaluated five model-selection procedures: RMSSE, ERA, OWA, CCG-AHSC, and CCG-AHSCD. These procedures were assessed under 91:9, 80:20, and 70:30 training–testing partitions and across forecasting horizons from $h = 1$ to $h = 12$. The global evaluation encompassed 2,203 Smooth, 1,880 Intermittent, 229 Erratic, and 1,021 Lumpy demand series.

### 4.1. Distribution of selected forecasting models

Figure 2 displays the overall frequency at which each forecasting model was selected by the respective procedures and experimental partitions.

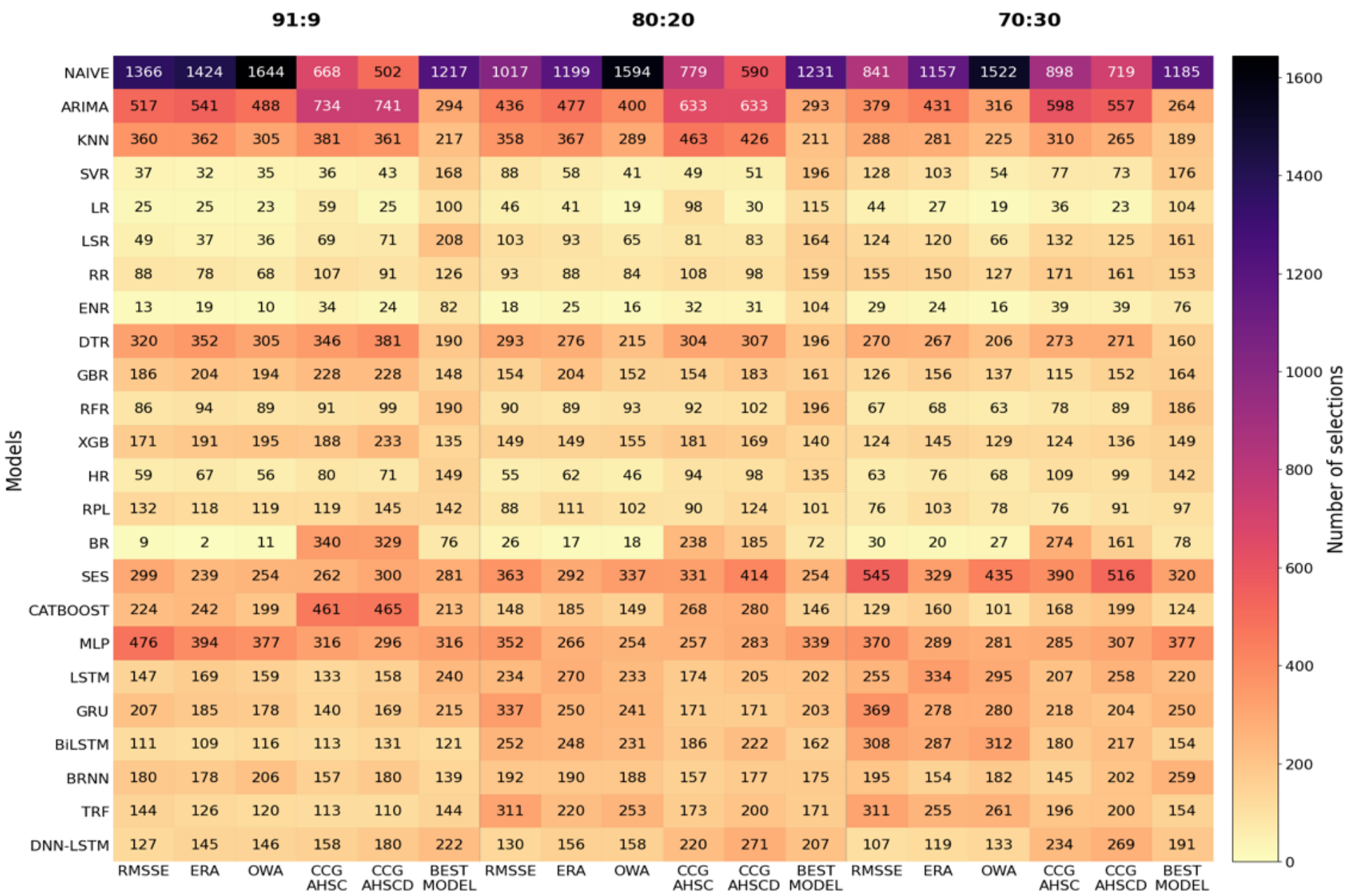

**Figure 2.** Global frequency of forecasting-model selection by selector and experimental partition.

Selections were distributed among several forecasting models and demonstrated variability across selectors and partitions. The BEST MODEL similarly displayed a distributed selection pattern, rather than focusing on a single forecasting model.

## 4.2. Selector performance by demand class

Figures 3–6 illustrate the progression of median GRA across forecasting horizons for each demand class and experimental partition. The BEST MODEL serves solely as an ex post reference, while the comparative analysis focuses on RMSSE, ERA, OWA, CCG-AHSC, and CCG-AHSCD.

### *4.2.1. Smooth demand*

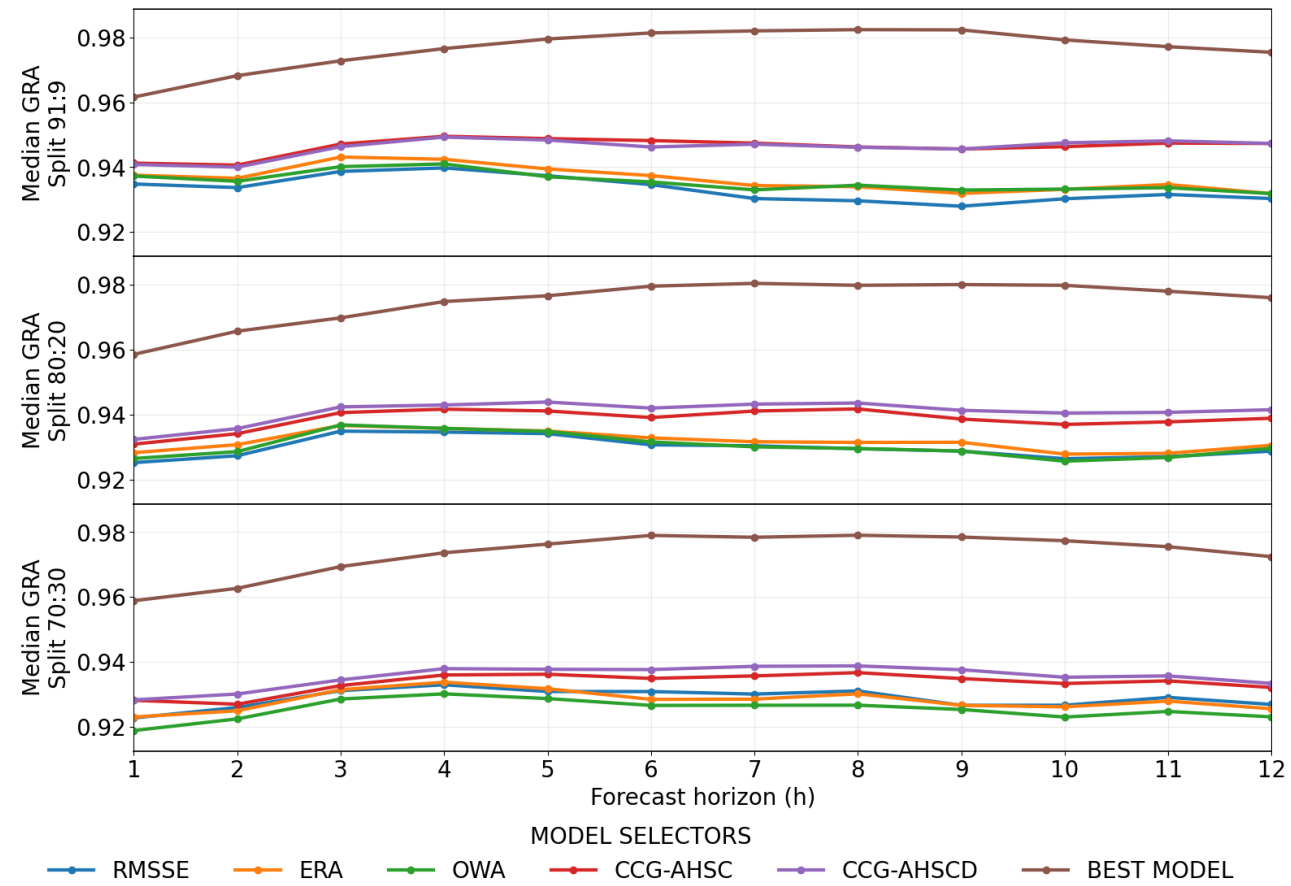

**Figure 3.** Evolution of median GRA by forecasting horizon and partition for Smooth demand series

The smooth series demonstrated stable GRA profiles throughout the forecasting horizon. For the 91:9 partition, CCG-AHSC achieved the highest median GRA (0.9463), closely followed by CCG-AHSCD (0.9461). In the 80:20 and 70:30 partitions, CCG-AHSCD ranked first, with median GRA values of 0.9408 and 0.9355, respectively. The global Friedman tests revealed statistically significant differences among selectors in all three partitions ($p < 0.001$). However, for the 91:9 partition, the post-hoc comparison between CCG-AHSC and CCG-AHSCD was not statistically significant after Holm correction. In contrast, for the 80:20 and 70:30 partitions, CCG-AHSCD consistently ranked first across all 12 forecasting horizons.

#### *4.2.2. Intermittent demand*

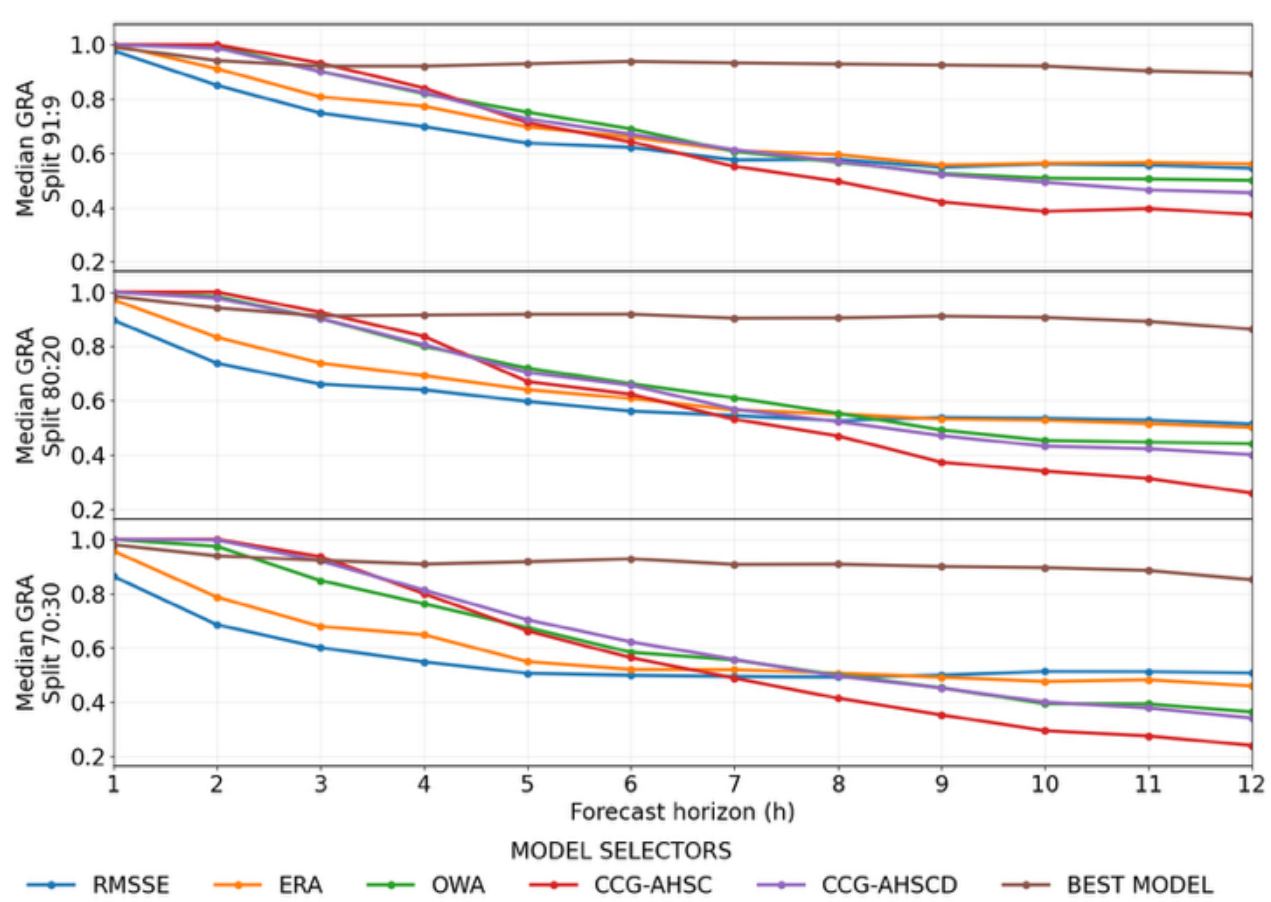

**Figure 4.** Evolution of median GRA by forecasting horizon and partition for Intermittent demand series

For intermittent series, OWA achieved the highest median GRA under the 91:9 partition (0.6975) and the 80:20 partition (0.6720). In contrast, CCG-AHSCD attained the highest median GRA under the 70:30 partition (0.6404). The global Friedman tests did not reveal statistically significant differences among the five selectors in any of the three partitions ($p > 0.05$). Consequently, OWA and CCG-AHSCD recorded the highest descriptive median GRA values for their respective partitions, although no statistically significant global differences were observed among selectors. Additionally, the GRA profiles exhibited a progressive decrease as the forecasting horizon increased, particularly under the 80:20 and 70:30 partitions.

#### *4.2.3. Erratic demand*

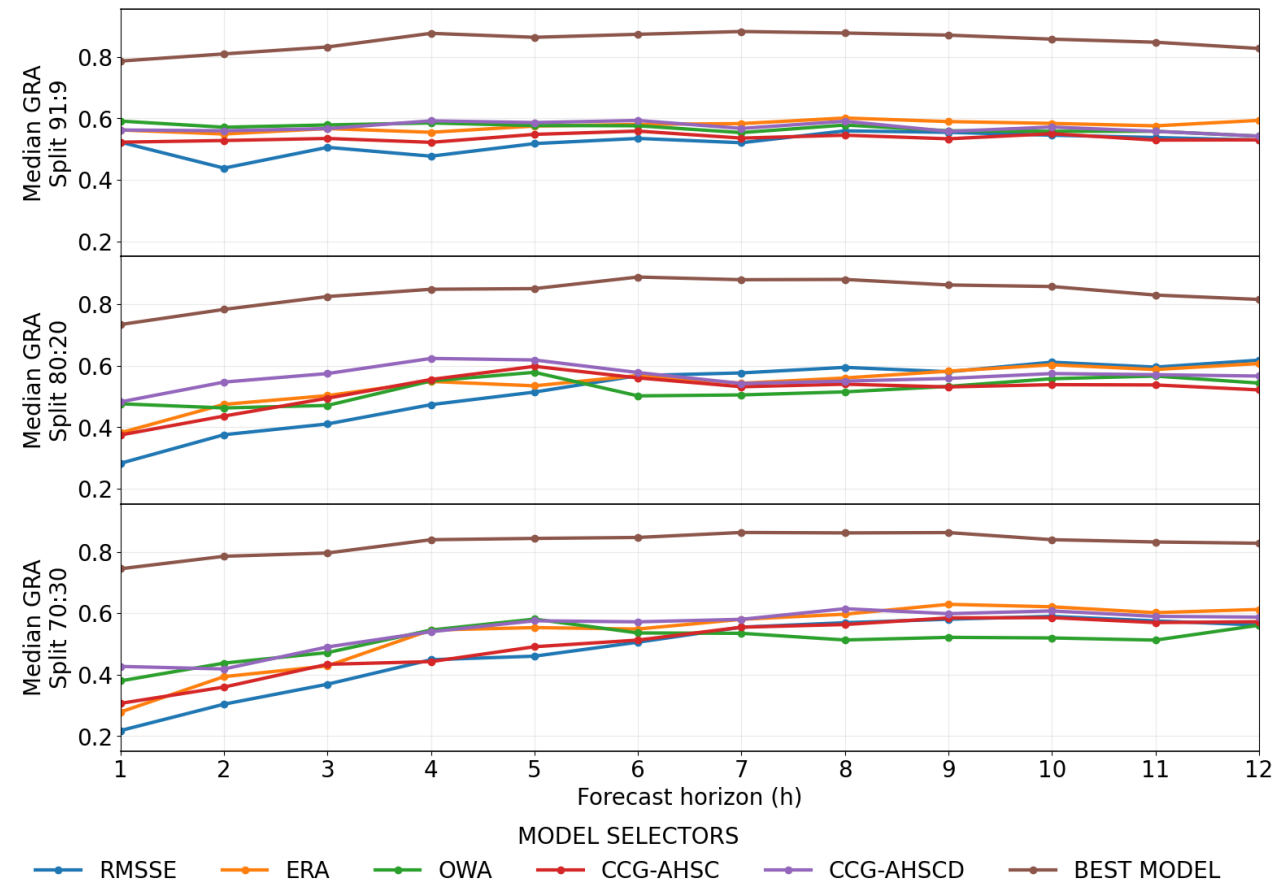

**Figure 5.** Evolution of median GRA by forecasting horizon and partition for Erratic demand series

In the Erratic series, ERA achieved the highest median GRA under the 91:9 partition, recording a value of 0.5765, followed by CCG-AHSCD at 0.5709. For the 80:20 and 70:30 partitions, CCG-AHSCD ranked first, with median GRA values of 0.5658 and 0.5500, respectively. The Friedman tests revealed statistically significant global differences across all three configurations: 91:9 ($p = 1.45 \times 10^{-7}$), 80:20 ($p = 0.0068$), and 70:30 ($p = 2.86 \times 10^{-4}$).

### *4.2.4. Lumpy demand*

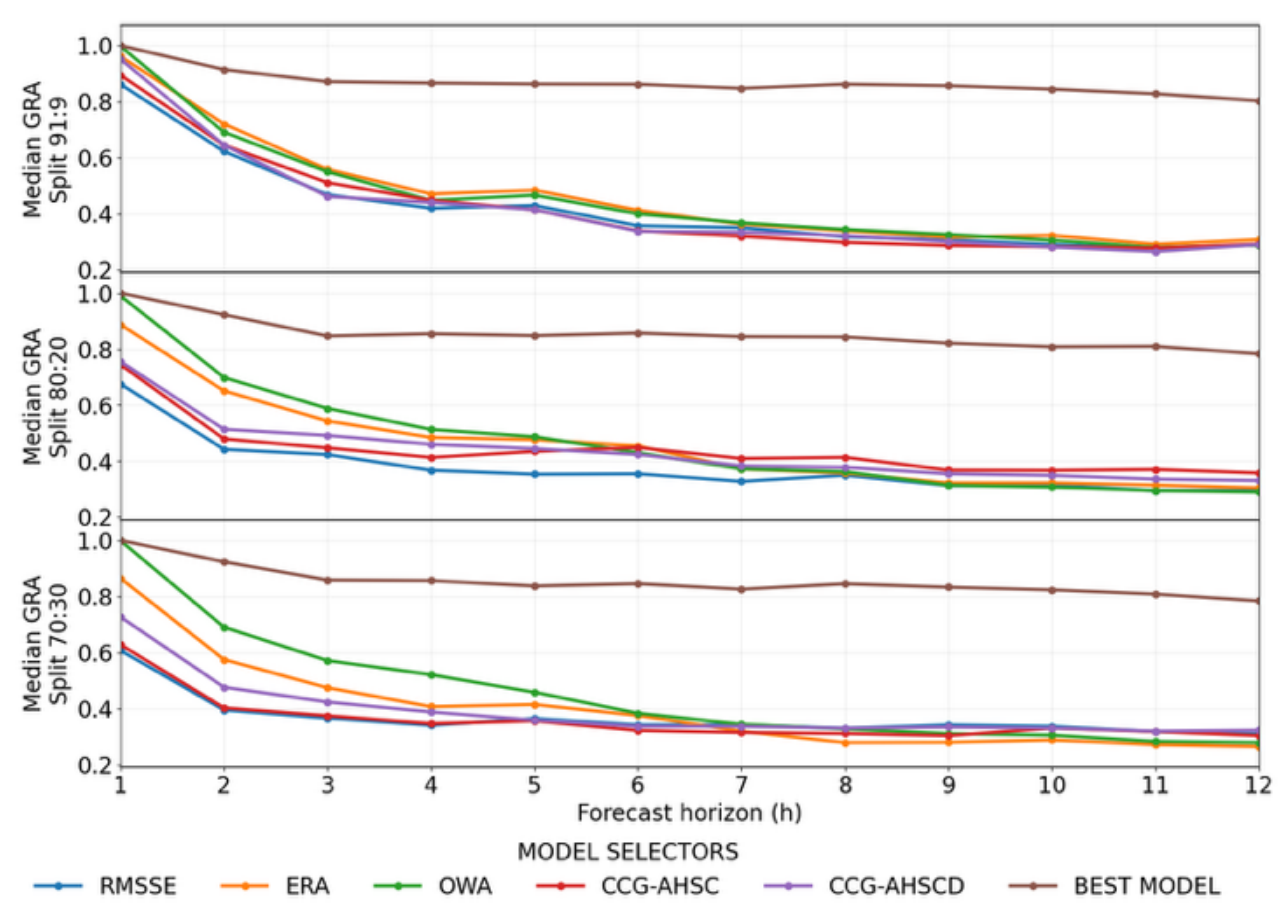

**Figure 6.** Evolution of median GRA by forecasting horizon and partition for Lumpy demand series

Within the Lumpy series, ERA achieved the highest median GRA under the 91:9 partition, recording a value of 0.4625. OWA ranked first under the 80:20 and 70:30 partitions, with median GRA values of 0.4706 and 0.4557, respectively. Statistically significant global differences among selectors were observed across all three partitions: 91:9 ($p = 5.05 \times 10^{-6}$), 80:20 ($p = 5.35 \times 10^{-4}$), and 70:30 ($p = 0.0411$). Additionally, the GRA profiles demonstrated a pronounced decline as the forecasting horizon increased.

## 4.3. Summary of global results

Table 3 presents the global results categorized by demand class and experimental partition. It includes the selector with the highest median GRA, the second-ranked selector, and the results of the global Friedman test.

**Table 3.** Summary of global selector performance by demand class and experimental partition.

| Demand class | Split | Best selector | Median GRA | Second selector | Median GRA | Friedman p-value | Statistical result |
|---|---|---|---|---|---|---|---|
| Smooth | 91:9 | CCG-AHSC | 0.9463 | CCG-AHSCD | 0.9461 | $7.07 \times 10^{-9}$ | Significant |
| Smooth | 80:20 | CCG-AHSCD | 0.9408 | CCG-AHSC | 0.9385 | $5.13 \times 10^{-9}$ | Significant |
| Smooth | 70:30 | CCG-AHSCD | 0.9355 | CCG-AHSC | 0.9335 | $2.53 \times 10^{-9}$ | Significant |
| Intermittent | 91:9 | OWA | 0.6975 | ERA | 0.6918 | 0.1478 | Not significant |
| Intermittent | 80:20 | OWA | 0.6720 | CCG-AHSCD | 0.6552 | 0.2720 | Not significant |
| Intermittent | 70:30 | CCG-AHSCD | 0.6404 | OWA | 0.6254 | 0.2720 | Not significant |
| Erratic | 91:9 | ERA | 0.5765 | CCG-AHSCD | 0.5709 | $1.45 \times 10^{-7}$ | Significant |
| Erratic | 80:20 | CCG-AHSCD | 0.5658 | ERA | 0.5415 | 0.0068 | Significant |
| Erratic | 70:30 | CCG-AHSCD | 0.5500 | ERA | 0.5323 | $2.86 \times 10^{-4}$ | Significant |
| Lumpy | 91:9 | ERA | 0.4625 | OWA | 0.4554 | $5.05 \times 10^{-6}$ | Significant |
| Lumpy | 80:20 | OWA | 0.4706 | ERA | 0.4575 | $5.35 \times 10^{-4}$ | Significant |
| Lumpy | 70:30 | OWA | 0.4557 | ERA | 0.4007 | 0.0411 | Significant |

The top-performing selector differed depending on the demand class and experimental partition. CCG-AHSC or CCG-AHSCD achieved the highest rank in all Smooth configurations and in two of the three Erratic configurations. OWA was ranked first in two Intermittent and two Lumpy configurations. ERA attained the highest rank for Erratic and Lumpy demand under the 91:9 partition.

## 5. Discussion

The results provide no evidence of a universally superior model-selection mechanism across heterogeneous demand conditions. Instead, selector performance varies according to demand structure, historical data availability, and forecasting horizon. CCG-AHSC and CCG-AHSCD were consistently competitive for Smooth demand, whereas OWA and ERA showed advantages in several Intermittent and Lumpy configurations. Erratic demand exhibited an intermediate behavior, with the preferred selector changing across training–testing partitions. These findings support the central premise of the study: the forecasting-model selection problem should not be reduced to the application of a single invariant criterion, because the effectiveness of the selection rule itself is context dependent. This interpretation is consistent with previous forecasting research showing that model performance varies with demand characteristics and evaluation conditions (Petropoulos et al., 2022; Syntetos et al., 2005; Syntetos & Boylan, 2005).

For Smooth demand, CCG-AHSC and CCG-AHSCD achieved the strongest and most stable performance across the three training–testing partitions. The relatively regular structure of Smooth series may provide more consistent performance information for differentiating among candidate models, allowing class-conditioned multicriteria procedures to exploit several complementary criteria without excessive instability. In this setting, CCG first reduces the candidate space using class-specific evidence, after which AHSC or AHSCD applies Pareto-based and hierarchical decision rules. The persistence of their advantage across partitions suggests that, when demand behavior is comparatively stable, conditioning the selection process on demand structure can provide a more effective basis for model discrimination than applying the same criterion uniformly to all series. The small difference between CCG-AHSC and CCG-AHSCD under the 91:9 configuration should nevertheless be interpreted cautiously, because their post-hoc difference was not statistically significant.

Intermittent demand produced a substantially different result. OWA attained the highest descriptive GRA under the 91:9 and 80:20 partitions, while CCG-AHSCD ranked first under 70:30; however, the Friedman tests did not identify statistically significant differences among selectors in any of these configurations. This lack of statistical separation suggests that the sparse occurrence of positive demand and the prevalence of zero-demand periods may limit the ability of any single selection mechanism to establish a consistently superior model assignment. Under these conditions, relatively simple benchmark-relative rules such as OWA may remain competitive with more elaborate class-conditioned procedures. Accordingly, the observed ranking differences for Intermittent demand should be interpreted as evidence of contextual variation rather than as proof of selector superiority.

Erratic and Lumpy demand further demonstrate the importance of historical data availability. For Erratic demand, ERA ranked first under the 91:9 partition, whereas CCG-AHSCD achieved the highest performance under 80:20 and 70:30. For Lumpy demand, ERA led under 91:9, while OWA ranked first under the two partitions containing less training information. In contrast to the Intermittent case, these differences were statistically significant. The change in the preferred selector as the training proportion decreases indicates that selector effectiveness is sensitive not only to demand structure but also to the amount of historical information available for estimating models and characterizing their previous performance. Consequently, a selection rule that performs well when relatively long histories are available cannot be assumed to retain the same advantage when the information used for model estimation and discrimination becomes more limited.

Forecast horizon introduces an additional source of variation. GRA generally declined as the horizon increased, particularly for Intermittent and Lumpy demand. More importantly, the trajectories indicate that relative selector performance is not necessarily constant throughout the complete forecasting horizon. This finding has practical implications for multi-period planning, because a selector identified as adequate from an aggregate evaluation may not be equally effective at short and longer horizons. Selector assessment should therefore consider both overall performance and its evolution with $h$, rather than treating the forecasting horizon as a fixed background characteristic. In this sense, horizon length becomes part of the selection context together with demand class and historical data availability.

A further contribution emerges from comparison with the ex post BEST MODEL reference. For every evaluated SKU, at least one optimized candidate model provides the highest attainable performance within the available model set, but the identity of that model is distributed across forecasting methods rather than being concentrated in a single universally dominant model. This result shifts the interpretation of the forecasting problem. The main limitation is not necessarily the absence of capable forecasting models; high-performing alternatives are already present within the candidate pool. Rather, the unresolved problem is the ability to identify ex ante which available model will be most appropriate for a given series and forecasting context before future demand is observed. The gap between the model selected in advance and the best model observable ex post therefore provides a direct empirical indication of the remaining opportunity for improving automatic model-selection mechanisms.

This distinction also clarifies the role of the proposed class-conditioned pipelines. CCG-AHSC and CCG-AHSCD should not be interpreted as universally dominant replacements for existing selectors. Their value lies instead in demonstrating that selection rules can explicitly incorporate demand structure before the final model decision is made. The empirical results show that this conditioning is particularly effective for Smooth demand and in several Erratic configurations, but that simpler mechanisms remain competitive or preferable under other conditions. Such behavior indicates that conditioning the model-selection rule on demand structure can be beneficial, but that the advantage of such conditioning is itself context dependent.

Taken together, these findings extend recent evidence on context-dependent forecasting-model suitability to the model-selection layer. Reina-Jiménez et al. (2026) showed that structural time-series characteristics can be exploited through an interpretable meta-learning framework to recommend suitable forecasting algorithms, reinforcing the broader finding that no single forecasting model is universally appropriate. The present study identifies an analogous phenomenon at a higher decision level: the mechanism used to choose among candidate forecasting models is itself not universally dominant. Selector suitability varies with demand pattern, historical data availability, and forecasting horizon, supporting the view that the selector should be treated as a distinct component of the forecasting decision. This distinction is important because adapting the forecasting model and adapting the rule used to select that model represent related but different decision problems. The resulting conditioned perspective provides a basis for future mechanisms aimed at reducing the gap between ex ante model assignment and the best attainable model identified ex post.

## 6. Conclusion

The results indicate that none of the five evaluated mechanisms consistently outperforms the others across all demand classes, training–testing partitions, and forecasting horizons. CCG-AHSC and CCG-AHSCD demonstrated greater competitiveness for Smooth demand and in several Erratic configurations, while OWA and ERA exhibited advantages under specific Intermittent and Lumpy conditions.

These findings support a conditional perspective on the model-selection problem, in which the selector itself forms part of the forecasting decision and its effectiveness depends on demand structure, historical data availability, and forecasting horizon. In this context, CCG-AHSC and CCG-AHSCD provide adaptive strategies by initially narrowing the candidate model space according to demand class and subsequently applying class-specific multicriteria decision rules.

The application of GRA supplemented conventional forecasting metrics by introducing a measure of volumetric coherence relevant to purchasing, replenishment, and inventory decisions. Furthermore, the ex post best-model reference established an empirical upper bound for evaluating how closely each selector approximates the best available alternative within the optimized candidate set.

The identification of an ex post best model for each SKU demonstrates that the evaluated model space includes alternatives capable of achieving high performance. The primary challenge remains to consistently identify these models in advance, without access to future demand. This finding highlights a research direction focused on narrowing the gap between ex ante selection and the best model observable ex post. Future research should aim to enhance this capability by explicitly integrating demand class, forecasting horizon, and historical model-performance dynamics into the selection rule. Future research should also examine the sensitivity of the CCG framework to its class-specific weighting coefficients and candidate-retention values in order to assess the robustness of the selection results under alternative parameter configurations. Additionally, these mechanisms should be evaluated using direct operational criteria, such as inventory costs, shortages, and service levels, and their generalizability should be validated across diverse datasets and demand frequencies.

**Author contributions statement**
Adolfo González: Conceptualization, methodology, software, validation, formal analysis, investigation, data curation, writing – original draft, and visualization. Víctor Parada: Supervision and manuscript review. Both authors reviewed and approved the final version of the manuscript.

**Disclosure statement**
The authors declare that they have no known competing financial interests or personal relationships that could have appeared to influence the work reported in this paper.

**Funding**
This research received no external funding.

**Data availability statement**
The datasets analyzed in this study are publicly available. The Walmart source is available via Kaggle (Yasser, 2021): https://www.kaggle.com/code/yasserh/walmart-sales-prediction-best-ml-algorithms. The M3 (Makridakis & Hibon, 2000) data are available from the International Institute of Forecasters repository: https://forecasters.org/resources/time-series-data/m3-competition/. The M4 (Makridakis et al., 2018) data are available from the M4 Competition repository: https://www.m4.unic.ac.cy/the-dataset/. The M5 data are available from the Kaggle M5 (Makridakis, Spiliotis, et al., 2022) Forecasting Accuracy competition: https://www.kaggle.com/c/m5-forecasting-accuracy. The BRAF and MAN datasets are publicly available through the *Spare-Part-Demand-Forecasting* repositories used for benchmarking intermittent spare-parts demand forecasting methods (de Haan, 2021a, 2021b): https://github.com/danieldehaan96/spdf and https://github.com/KhueNguyenTK/Spare-Part-Demand-Forecasting. These repositories provide the industrial spare-parts datasets employed in this study, including the BRAF and MAN series. The Beer, Analgesics, and Cheeses datasets are publicly available as part of the Dominick's Dataset maintained by the Kilts Center for Marketing at The University of Chicago Booth School of Business (Kilts Center for Marketing, n.d.):
https://www.chicagobooth.edu/research/kilts/research-data/dominicks. Specifically, the Beer data were obtained from the *wber* movement file, the Analgesics data from the *wana* movement file, and the Cheeses data from the *wche* movement file. For the present study, sales were aggregated across stores for each unique UPC, thereby producing one weekly demand time series per SKU and removing the store dimension while preserving the temporal demand structure.

## Appendix A

**Table A1.** Candidate forecasting models and optimization groups.

| Model | Cluster | Explanation |
|---|---|---|
| NAÏVE | - | Naïve forecasting is a simple benchmark method that uses the most recent observed value as the forecast for future periods. It provides a basic reference against which the performance of more sophisticated forecasting models can be evaluated (Makridakis & Hibon, 2000). |
| ARIMA | ES | Autoregressive Integrated Moving Average is a classical time series approach that captures linear relationships through autoregressive components, moving averages, and differencing to address trends. It is particularly effective for stationary data or for series that can be rendered stationary through appropriate transformations (Box et al., 2015). |
| KNN | ES | K-Nearest Neighbors is a nonparametric technique that estimates values based on the average of the k nearest neighbors in the feature space. It is characterized by its simplicity, interpretability, and its capacity to handle nonlinear data structures (Altman, 1992). |
| DTR | ES | Decision Tree Regression generates partitions through hierarchical decision rules, resulting in interpretable models, although they are prone to overfitting (Breiman et al., 2017). |
| RFR | ES | Random Forest Regressor constructs an ensemble of multiple decision trees trained on random subsets of the data, thereby improving predictive accuracy and reducing overfitting (Breiman, 2001). |
| PLR | ES | Polynomial Regression extends linear regression by incorporating polynomial terms, thereby enabling the modeling of nonlinear relationships in a straightforward manner (Montgomery et al., 2021). |
| MLP | ES | Multilayer Perceptron is an artificial neural network with one or more hidden layers, capable of capturing complex nonlinear relationships and widely used in regression and classification tasks (Haykin, 1994). |
| LR | - | Linear Regression is a fundamental regression model that estimates the relationship between independent and dependent variables by minimizing the squared error (Montgomery et al., 2021). |
| SVR | SCS | Support Vector Regression is an extension of support vector machines for regression tasks, designed to identify a function that lies within a specified tolerance margin from the observed values, employing kernel functions to capture complex relationships (Smola & Schölkopf, 2004). |
| LSR | SCS | Lasso Regression incorporates an L1 regularization penalty, thereby facilitating automatic variable selection by shrinking less relevant coefficients to zero (Tibshirani, 1996). |
| RR | SCS | Ridge Regression applies an L2 regularization penalty, which mitigates multicollinearity without eliminating variables (Hoerl & Kennard, 1970). |
| ENR | SCS | Elastic Net Regression combines L1 and L2 penalties, enabling the selection of relevant variables while controlling for collinearity (Zou & Hastie, 2005). |
| GBR | SCS | Gradient Boosting Regressor constructs decision trees sequentially, where each new tree corrects the errors of the preceding ones, achieving high predictive performance at the cost of increased calibration complexity (Friedman, 2001). |
| XGBoost | SCS | An optimized implementation of gradient boosting improves computational speed, regularization, and the handling of missing data, making it particularly effective for structured prediction problems (Chen & Guestrin, 2016). |
| HR | SCS | Huber Regressor is a robust regression technique that employs the Huber loss function, combining sensitivity to small squared errors with resistance to outliers (Huber, 1964). |
| BR | SCS | Bayesian Ridge Regression introduces prior distributions over the regression coefficients, thereby incorporating uncertainty into the parameter estimates (MacKay, 1992). |
| SES | SCS | Simple Exponential Smoothing is applied to time series without trend or seasonality, assigning greater weight to recent observations through an exponential decay rule (Brown, 1959). |
| CatBoost | SCS | A boosting algorithm that efficiently handles categorical variables without explicit encoding and reduces overfitting through ordered boosting (Prokhorenkova et al., 2018). |
| LSTM | SCS | Long Short-Term Memory is a recurrent architecture designed to capture long-term dependencies in sequential data, particularly useful in time series applications (Hochreiter & Schmidhuber, 1997). In the present study, a regression-oriented variant is implemented, consisting of two LSTM layers (with 128 and 64 units, respectively, and tanh activation), a dropout mechanism of 30%, an intermediate dense layer with 32 neurons (ReLU activation), and a linear output layer. The model is compiled using the Adam optimizer and employs mean squared error (MSE) as the loss function. |
| GRU | SCS | Gated Recurrent Unit is a recurrent architecture designed to model temporal dependencies through a simplified gating mechanism, reducing computational complexity compared to LSTM while maintaining an adequate capacity to represent sequential dynamics, which makes it particularly suitable for time series applications (Cho et al., 2014). In the present study, a regression-oriented variant is implemented, consisting of a single GRU layer with 64 units and tanh activation, followed by an intermediate dense layer with 64 neurons and ReLU activation. The output layer corresponds to a dense layer whose dimensionality is adjusted according to the prediction horizon under consideration, thereby enabling direct multi-step forecasting. The model is compiled using the Adam optimizer and employs mean squared error (MSE) as the loss function, ensuring consistency in predictive performance evaluation. |
| BiLSTM | SCS | Bidirectional Long Short-Term Memory Neural Network is a bidirectional recurrent architecture that extends the LSTM model by simultaneously processing the temporal sequence in both forward and backward directions, thereby enabling the capture of past and future dependencies within the observation window and improving the representation of complex temporal patterns (Graves & Schmidhuber, 2005; Schuster & Paliwal, 1997). In the present study, a multi-step regression-oriented variant is implemented, consisting of two stacked BiLSTM layers with 64 and 32 units, respectively. The first layer operates in full sequential mode, while the second condenses the temporal information into a final state representation. To mitigate the risk of overfitting, dropout regularization with a rate of 20% is applied after each recurrent layer. The resulting representation is subsequently processed through two intermediate dense layers with 64 and 32 neurons, respectively, both using ReLU activation, before applying a dense output layer with linear activation whose dimensionality is adjusted according to the prediction horizon under consideration, thereby enabling direct multi-step forecasting. The model is trained using the Adam optimizer and employs mean |
| | | squared error (MSE) as the loss function, ensuring consistency in predictive performance evaluation. |
| BRNN | SCS | Bidirectional Simple Recurrent Neural Network is a bidirectional recurrent architecture based on simple recurrent units, which processes the input temporal sequence simultaneously in forward and backward directions, thereby capturing bidirectional temporal dependencies within the observation window, albeit with a simpler structure than LSTM or GRU variants (Schuster & Paliwal, 1997). In the present study, a multi-step regression-oriented variant is implemented, consisting of a single bidirectional SimpleRNN layer with 64 units and tanh activation, which condenses the temporal information into a final representative state. The resulting representation is subsequently processed through an intermediate dense layer with 64 neurons and ReLU activation, before applying a dense output layer whose dimensionality is adjusted according to the prediction horizon under consideration, thereby enabling direct multi-step forecasting. The model is trained using the Adam optimizer and employs mean squared error (MSE) as the loss function, ensuring consistent evaluation of predictive performance. |
| TRF | SCS | Transformer Model for Time Series is an architecture based on autoregressive self-attention mechanisms that dispenses with explicit recurrence and enables the modeling of long-range temporal dependencies through multi-head attention, making it particularly suitable for capturing complex and non-local relationships in temporal sequences (Vaswani et al., 2017). In the present study, a multi-step regression-oriented variant is implemented, beginning with a dense projection of the input into a 64-dimensional representation space, followed by a Transformer block composed of a multi-head attention mechanism with two heads and an internal feed-forward network of 128 units. The block incorporates residual connections, layer normalization, and dropout regularization, allowing stable information propagation and robust learning of temporal dependencies. The resulting representation is subsequently transformed through flattening and two intermediate dense layers with 64 and 32 neurons, respectively, both using ReLU activation, before applying a dense output layer with linear activation whose dimensionality is adjusted according to the prediction horizon under consideration, thereby enabling direct multi-step forecasting. The model is trained using the Adam optimizer and employs mean squared error (MSE) as the loss function, ensuring consistency in predictive performance evaluation. |
| DNN-LSTM | SCS | Deep Neural Network–Long Short-Term Memory is a hybrid neural architecture that combines the temporal modeling capabilities of LSTM networks with fully connected deep layers for nonlinear feature transformation and regression. LSTM components are well suited to capturing long-term dependencies in sequential data (Hochreiter & Schmidhuber, 1997), while deep feedforward layers enhance the model's capacity to learn complex nonlinear representations (Goodfellow et al., 2016). In the present study, the model consists of two stacked LSTM layers with 128 and 64 units and tanh activation, each followed by 30% dropout. The recurrent representation is subsequently processed by dense layers with 128, 64, and 32 neurons using ReLU activation, with additional 20% dropout after the first two dense layers. The output layer is adjusted to the forecasting horizon, enabling direct multi-step prediction. The model is trained using the Adam optimizer and mean squared error (MSE) as the loss function. |

**Table A2.** Parameter ranges explored for the models considered in this study.

| Model | Parameters |
|---|---|
| NAÏVE | Defaults |
| ARIMA | p=0-5, d=0-3, q=0-5 |
| KNN | n_neighbors between 1 and 35 |
| DTR | max_depth between 1 and 32; min_samples_split = 2; min_samples_leaf = 1; random_state = 21 |
| RFR | n_estimators between 1 and 100; random_state = 21 |
| PLR | degree between 1 and 3; include_bias = False; interaction_only = True |
| MLP | hidden_layer_sizes = [layer 1: 16–32; layer 2: 0–32; layer 3: 0–32]; activation = 'relu'; solver = 'adam'; max_iter = 1000; early_stopping = True; n_iter_no_change = 10; random_state = 21 |
| LR | Defaults |
| SVR | kernel = 'rbf'; C between 0.01 and 20; degree between 2 and 5; epsilon between 0.01 and 1; gamma = 'scale'. |
| LSR | alpha between 0.01 and 5; random_state = seed_value; max_iter = 8000. |
| RR | alpha between 0.1 and 10; max_iter = 8000 |
| ENR | alpha between 0.01 and 0.02; l1_ratio between 1e-3 and 0.1 |
| GBR | n_estimators between 10 and 2000; learning_rate between 0.1 and 0.3; random_state = seed_value |
| XGBoost | n_estimators between 10 and 2000; learning_rate between 0.1 and 0.3; max_depth between 3 and 5; enable_categorical = True; random_state = seed_value |
| HR | epsilon between 1 and 3; max_iter = 1000; alpha = 0.0001; tol = 1e-4 |
| BR | max_iter between 50 and 1000; alpha_1 between 1e-6 and 1e-2; alpha_2 between 1e-6 and 1e-2 |
| SES | smoothing_level between 0.01 and 0.9; optimized = False |
| CatBoost | iterations between 100 and 1000; depth between 3 and 10; learning_rate between 0.01 and 0.3; loss_function = 'Huber:delta=1.0'; train_dir = tmp_dir; verbose = False; random_seed = 21 |
| LSTM | epochs between 1 and 80 |
| GRU | epochs between 1 and 80 |
| BiLSTM | epochs between 1 and 80 |
| BRNN | epochs between 1 and 80 |
| TRF | epochs between 1 and 80 |
| CNN-LSTM | epochs between 1 and 80 |